# The Newton Scheme for Deep Learning

Junqing Qiu, Guoren Zhong, Yihua Lu, Kun Xin, Huihuan Qian, Xi Zhu*

School of Science and Engineering,

The Chinese University of Hong Kong, Shenzhen.

Shenzhen, Guangdong, 518172, China

zhuxi@cuhk.edu.cn

Abstract

We introduce a neural network (NN) strictly governed by Newton's Law (I, II and III), with the nature required basis functions derived from the fundamental classic mechanics. Then, by classifying the training model as a quick procedure of "force pattern" recognition, we developed the Newton physics-based NS scheme. Once the force pattern is confirmed, the neuro network simply does the checking of the "pattern stability" instead of the continuous fitting by computational resource consuming big data-driven processing. In the given physics' law system, once the field is confirmed, the mathematics bases for the force field description actually are not diverged but denumerable, which can save the function representations from the exhaustless available mathematics bases. In this work, we endorsed Newton's Law into the deep learning technology and proposed Newton Scheme (NS). Under NS, the user first identifies the path pattern, like the constant acceleration movement. The object recognition technology first loads mass information, then, the NS finds the matched physical pattern and describe and predict the trajectory of the movements with nearly zero error. We compare the major contribution of this NS with the TCN, GRU and other physics inspired FIND-PDE methods to demonstrate fundamental and extended applications of how the NS works for the free-falling, pendulum and curve soccer balls. The NS methodology provides more opportunity for the future deep learning advances.

**Introduction**

The deep learning technology has attracted tremendous attention nowadays, and it is performing revolutionary changes across diverse social respects[1-3]. However, despite the wide successful application of the deep learning frameworks, like many other AI technologies, the robustness and the guarantees (error control) are naively data-driven, some schemes even require exascale data acquisition and processing[4], like face recognition. Some works even need 1.5B parameters to approach the infinite close but never reach the perfect match. The development trend of the huge data-driven AI technology pushes this new field towards the opposite extreme of Physics, which can use a single physics quantity, like gravity, to resolve a whole host of questions in planetary astronomy[5]. Split the difference and take the advantages between the data-driven and physics-driven methodologies, there could be an alternative more optimized way out for the solutions of the old questions. Previous works reported that physics inspired deep learning frameworks to discover the partial differential equations in nonlinear dynamical system[6-9] and applications in physics devices[10]. The previous works focused on the solution of partial differential equations (PDE) by polynomial and some trigonometrical function bases (FIND-PDE)[6], and it is still difficult to eliminate the systematic error due to limit intrinsic characters of polynomial functions. Here in this work, we endorse the Newton physics to guide the deep learning frameworks, purchasing the principle precision and computational efficiency. Combined with the computer vision object recognition input, this scheme provides a systematic error-free scenario for the prediction of object trajectories by tracking the force equations on the given objects.

**Method**

The vital step towards the solution of a specific classic physics (Newton physics) problem is to find the equation of relationship between the force and time, which means to exactly solve the equation $F = g(t)$. In fact, this can also evaluate the equation $x = g(t)$ and $v = g(t)$ as long as these equations are in the space formed by mathematic bases. To solve this Newton physics problem in NNs, we first take the typical MLP (Multilayer perceptron)[11] methodology for example. This is to use NNs (neural networks) to depict a function x=f(t), which can only accomplish the historical fitting part other than the forecast future movement. However, divide the MLP structure into two parts as the input/output layers connected by the hidden layers as

$$[f1(t) \quad f2(t) \quad f3(t) \quad ... \quad fn(t)] * [w1 \quad w2 \quad w3 \quad ... \quad wn],$$

here [f1(t) f2(t) f3(t) ... fn(t)] is the values in the last layer, the MLP network can provide the prediction ability through the polynomial function with activation function, as shown in Figure 1, the activation function can introduce the nonlinear correlations. Other typical NNs architecture like TCN(Temporal Convolutional Networks)[12], GRU(Gated Recurrent Units)[13] shares the same linear basis feature, similar to the MLP methodology, because the architecture is strongly restricted by the polynomial calculation rules, this architecture is inadequate for fitting the universal cases.

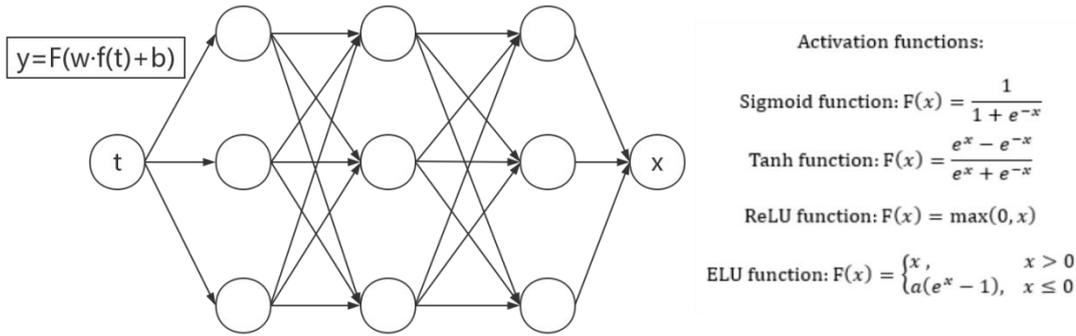

Figure 1: Scheme for the architecture of typical MLP NNs. F(x) is an activation function for calculation propagations in the NNs. Several common activation function like Sigmoid function, Tanh function, ReLU and ELU function is listed for example.

There are some other architecture such as recurrence can be applied for the prediction of classic movement. The well-developed NNs for the multivariate time series is GRU(Gated Recurrent Units)[13], and the TCN(Temporal Convolutional Networks)[12] is also proved to be powerful in this disciplines. Basically the above methods can provide numerical solutions to both of the fitting and prediction problem, however, once dealing with the large size objects or long-time trajectories, the accumulated error can be of huge problem.[14] Maziar et al proposed a Physics Informed Deep Learning (PIDL)[15] and Deep Hidden Physics Models (DHPM) [16] based on the training of the solution for partial differential equations. These methods provide further steps on the building of physics based deep learning architecture. Kutz et al [6-8] provided a regression strategy as "FIND-PDE" subjected to y=θ(t)x with both polynomial and trigonometric function for the PDE solution, it can solve various types of PDE shaped physics problems like burgers equation etc.[6] However, some basis in Kutz's basis are still unphysical likes $x^5$, which is still numerically polynomial and doesn't exist in nature. Moreover, the previous works neglects the object mass , which is also a critical quantity for the Newton physics.

The analytic or numerical solution to the physical equation tells the difference of understanding in a physical or mathematic way. Each force type corresponds to unique mathematic bases. For example, for the gravity, the displacement and time is correlated as the second order polynomial equations $x(t) = x(0) + at^2/2$, there is no other solutions with other math basis for the gravity in classic physics, thus the analytic approach to real objects' physical movement is significantly essential for training the AI or NNs the physics knowledge from Galileo, Newton or Einstein. The polynomial basis can only analytically or exactly represent the same polynomial serials $\sum_{i=0}^{\infty} a_i x^i$. Thus, it is analytically impossible and numerical inefficient for the polynomial basis to represent the other serials like $\sum_{i=0}^{\infty} \cos x_i$, $\sum_{i=0}^{\infty} e^{\omega i x}$, $\sum_{i=0}^{\infty} ln(x_i)$ etc. The main numerical errors rise from the fitting between two unmatched bases, like using $\sum_{i=0}^{\infty} a_i x^i$ to fit the periodic function $\sum_{i=0}^{\infty} \cos x_i$. However, the aforementioned functions correspond real physical pictures and are essential in both static and dynamic system, like harmonic oscillators ($\cos x_i$), damping oscillators ($e^{\omega i x}$), center force ($ln(x_i)$). The physical law provides the exclusions for the unreasonable mathematics description for the object movements.

Thus, to overcome the weakness in training the physics pictures, according to the feasible model types with the coverage of classic mechanics, in the NS we reconstruct new basis functions over the traditional polynomial one under the new scheme, which changes the previous fitting problem into a regression problem. The possible movement types in classic mechanics are enumerated in supplementary S1. According to Newton laws, most of natural movement is governed by some fundamental force with corresponding math basis patterns; we can assume the movement patterns are linearly combination of the individual movement patterns. $F = \sum_0^n \alpha_i * f_i$ and the total displacement can be defined as the same way, i.e.,

$$x = \int_0^t v_t dt = \int_0^t \int_0^t \sum_0^n \frac{f_n}{m} dt dt \quad (1)$$

The X matrix proposed here are named as Xu- matrix, which is a representation of the basis function for the typical force/movement-time patterns in classic physics as shown in supplementary S1. Currently, we already have 10 bases which are enough to fit most movement, the up limit of the complexity is $O(2^n)$ (n is number of bases). due to the large amount of force basis in the classic mechanics , we use attention mask [17-18] (in supplementary S6) ranked by the universality/ simplicity of the force types, as shown in supplementary S1.

Figure 2 shows the working flowchart of the NS, first the mass density of the object is recognized through the CV technology (the details are shown in supplementary S2). The NS first read in a short time of object trajectory, as colored by the green in Figure 2 (a), and then NS finish the regression by identifying the force types through the Xu-matrix as shown in Figure2 (b). Due to the different time dependent character of all the basis, it is strictly impossible to get the bijective function to reach the extremely low convergence, for a curve with more exponential character, if fitted by logarithmic or polynomial basis, the tight convergence criteria ($10^{-8}$) will numerical exclude the approaching solution. Once the force equations is identified, the whole system is just a mono function of time t and any quantity like force, location, velocity, acceleration velocity are all known. This step in Figure 2 (b) is the core part of the NS. The output of the NS is the physics law assured force types, which can be applied to predict the real objects movements as indicated by the red line in the Figure 2 (c). Different from the previous real time data driven architecture, once the physics force type is confirmed, the NS only do the checking calculation in the predictive range, i.e., once the predictive value matches the real trajectory, the NS keeps the force type, once the mismatch happens, as marked by the yellow line in Figure 2 (c). the NS stops the checking procedure accordingly and arouse the evaluating procedure with another checking. The data-driven events take only very fractional part of the whole events, for example, 1s in the 7s event in Figure 2(c).

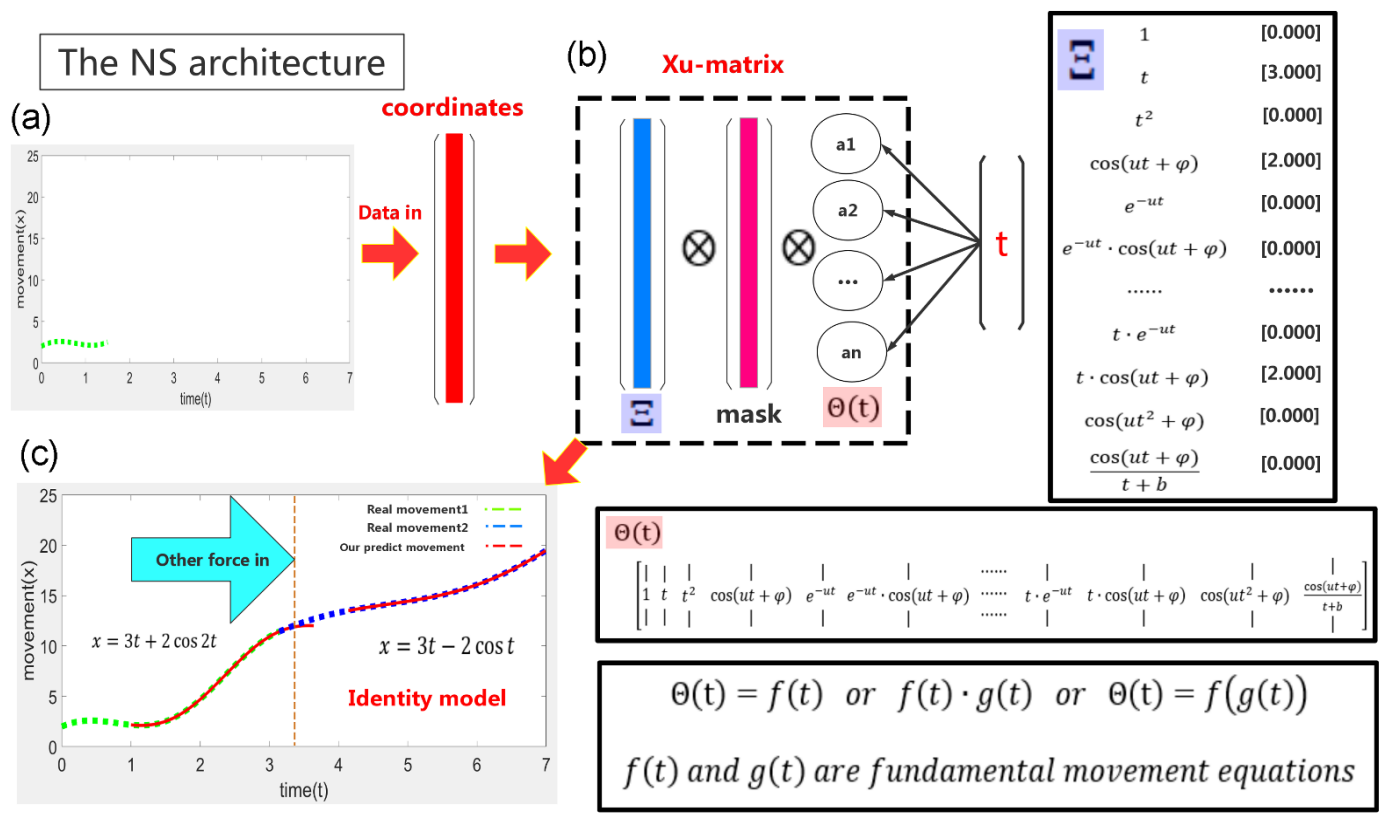

Figure 2. The flowchart of the working principle of NS. (a), the data input to the NS. (b), the processing with Xu-matrix. (c), the data output from NS, compared with the real experimental ones. The yellow line indicates a second processing once the checking procedure returns mismatch.

**Results and discussion**

Next, we illustrate the applications of NS in the real classic physics pictures. Figure 3 shows the prediction results for the free-falling objects with mass 10 mg from 10 m in the air with the MLP, GRU, TCN, FIND-PDE and NS scheme, the physics can be described as:

$$x(t) = x(0) + at^2/2 \quad \text{Eq (1)}.$$

We take the first 10s as the fitting area and the next 10s as the predicting time ranges. We can see the MLP, GRU and TCN methods all can well fit the training data with loss function less than $10^{-5}$, however, the MLP, TCN and GRU methods lost the prediction power for the future events, and the MLP methods only follow the linear trends for the historical information. All the above methods fail the prediction power due to the lack of the fundamental understanding about the physics. The FIND-PDE method and the NS methods can exactly predict the whole trajectories with limited training data, it takes no more than 10 seconds data to achieve the numerical convergence criterial of the $a$ value in Eq (1). One important conclusion can be derived is, the errors is magnified with time $t$ elapse for the inexact method like MLP, GRU or TCN, i.e. the near-perfect fit for the presence data makes no value for the real understanding at all.

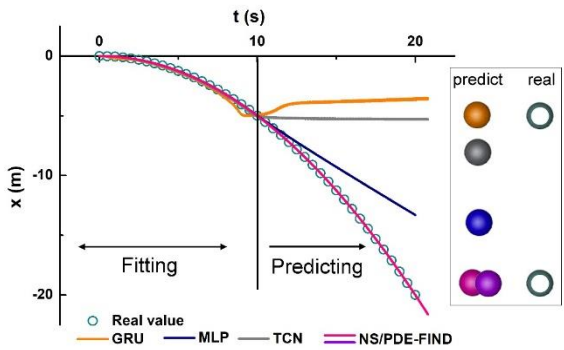

Figure 3: the trajectory prediction of the free-falling particle by MLP, GRU, TCN, FIND-PDE and NLS, the real values are benchmarked by the green points. The objects are perfect iron spheres with recognized mass density 7.9 $g/cm^3$. The color ball in the right part indicates the errors of $x(t)$ at t = 20s for each method.

The basic free-falling model depends on a simple polynomial time basis $t^2$; next we explore the non-polynomial object movements, the damping pendulum with the analytical mathematical solution

$$x(t) = ae^{-\gamma t} \cos(\omega t - \varphi) \quad \text{Eq (2)}$$

Here $a$ is the amplitude constant, $\gamma$ is the damping rate, $\omega$ is the angular frequency, $\varphi$ is the phase factor. Eq (2) is the typical damping function in solid state physics, especially in the light propagation in the interface of two dielectric materials[19]. The Xu-matrix basis for Eq(2) is $e^{-\gamma t} \cos(\omega t - \varphi)$ it is the combination products of $e^{-\gamma t}$ and $\cos(\omega t - \varphi)$, the NS methods can well fit the composite functions $f(t) * g(t)$ as long as the available bases are of the fundamental ones in Xu-matrix. The parameter $\omega$ and $\varphi$ are changeable parameters to get the lowest punishment value. Figure 4 (a) shows the trajectory-time dependent for the MLP, TCN, GRU, PDE-Find and the NS method. We can see the MLP, TCN and GRU schemes are still weak in the prediction and the PDE-FIND scheme also fails here due to the lack of capability of composite function. The NS method still can be robust for the damping pendulum picture as shown in the red and green circle in Figure 4 (a). Figure 4(b) shows the errors can be magnified for the inexact description of the damping pendulum system.

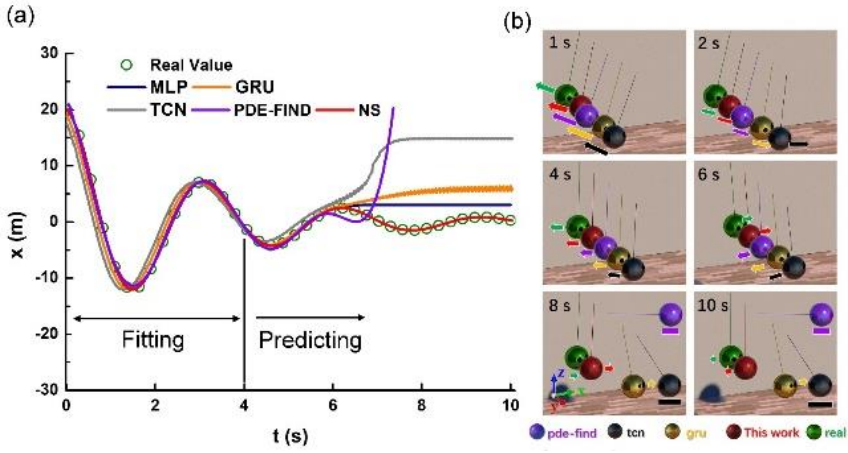

Figure 4: the trajectory prediction of the damping pendulum by MLP, GRU, TCN, FIND-PDE and NLS, the real values are benchmarked by the green points. The color ball in the right part indicates the errors of $x(t)$ at $t = 10s$ for each method.

The above two examples show the weakness of current neural network. First, it needs huge data which requires huge time and space cost, second and most important, as shown in Figure 3(b) and Figure 4(b), for the definition of "learning", a near-perfect method is not perfect at all, the "perfection" only makes sense when the neural network gets the ideas of physics law. Another advantage of the NS scheme is it doesn't need data normalization nor activation function. Activation function is used to make linear function nonlinear, but in physics, the total force can be linearly combined as the sum of each sub-force, $F = \sum_0^n f_i$, it manifests the nonlinear correlation itself, which naively replace the activation function. Moreover, the data normalization is no longer essential in NS because each physics law is independent and irrelevant, by passing the duplication of data processing from different physics is one of the most important features of NS.

**Applications**

The AI technology is widely applied in the sports field, such as table tennis[20], ice hockey [21] etc. The prediction of movements is crucial in many aspects, one of which is the training exercise in professional sports like table tennis. Take soccer for example,
by integrating the techniques and the desired ball trajectory, players can find the right way to score a goal. When a player hitting a ball, it is essential to forecast the ball's location, which is easy when the ball is not spinning. However, in most situations, the ball is flying with a self-spinning, resulting in a Magnus force on it, which makes it difficult to predict the position by traditional methods. Here we demonstrate that the NS scheme can work out to predict more complicated classic physics system, like magnus force in soccer. The curve-angle relationship can be described as below:[22]

$\theta(t) = \theta_0 + \frac{\lambda S}{\tau} t$    Eq (3)

$S(t) = \mathcal{L} \log\left(\frac{t}{\tau} + 1\right)$    Eq (4)

where  $\theta_0$  is the initial angle of velocity,
       $\lambda = 4C_n/C_D$,  $C_n$  and  $C_D$  are two parameters provided by experiments,
       $S = R\omega_0/v_{0,xy}$  is the spin parameter,
       $\mathcal{L}$  is the characteristic penetration length. The detail derivarions are in supplementary S3.
As shown in the inserted picture in Figure 5. The 2D curves can be described as the time dependent curve length S(t) and angle $\theta(t)$, as shown in Eq (3) and Eq (4), it is the basis of logarithm function. Consider the gravity into the 2D calculation, as shown in the red and black dash lines in Figure 5, the NS scheme can well desctibe the curve trajectory and predicts the acccurate final positions. By training the player the right position and time for the ball kicking , the NS scheme can be widely applied in the free ball kicking in the soccer sports.

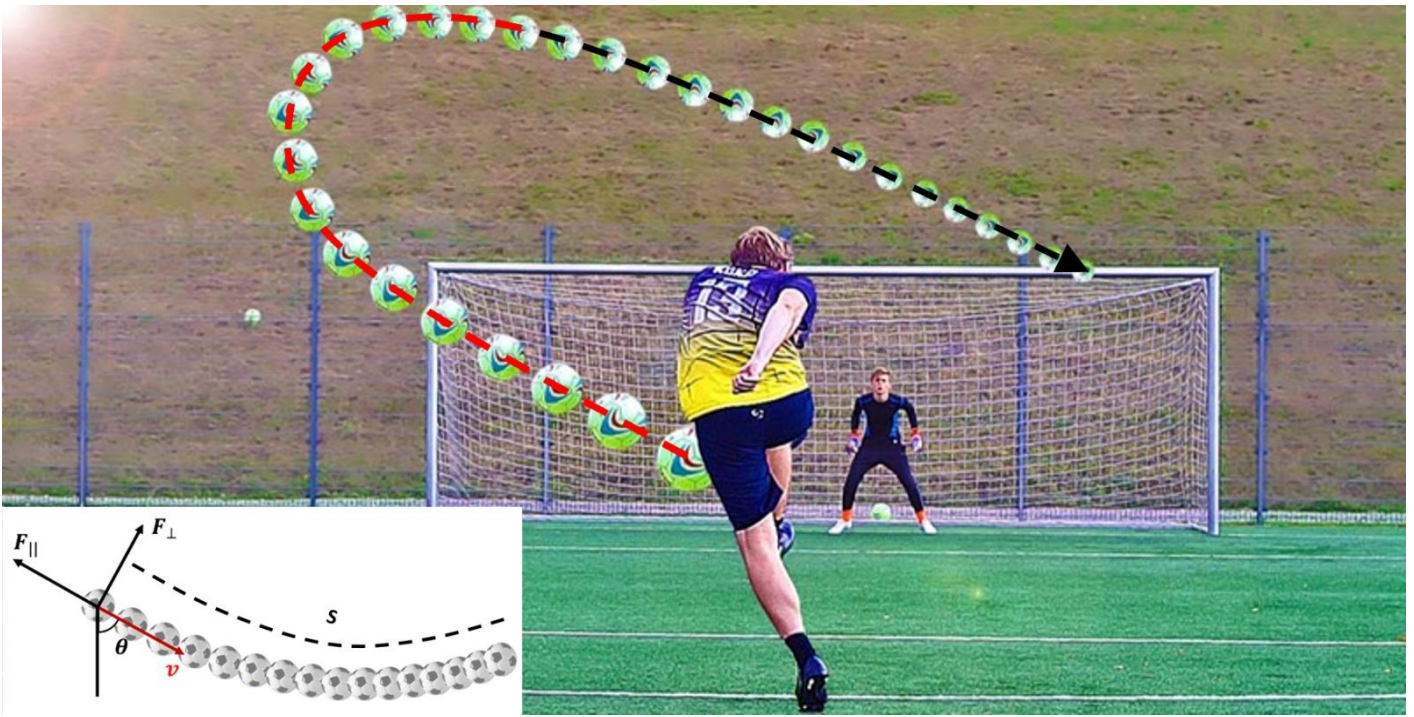

Figure 5: The trajectory of the curve ball in soccer. The red and black lines indicate the fitting and predicting procedures for the curve ball. The real trajectory is shown as the ball. The picture is taken from internet[23]. The inserted picture shows the force analysis for the running soccer ball in the 2D way measure by the time dependent curve length $S(t)$ and angle $\theta(t)$ respectively.

The trajectory of ball is in supplementary S5.

**Conclusion：**

In a conclusion, we build our NS based on MLP with knowledge about Newton law. The problem of nowadays neural network to handle with physic problem is can only handle problem with polynomial bases. However, the movements in world are governed by different types of forces; we can only depict only a few of them with polynomial bases. To solve this problem, we chose some fundamental bases in physical movement as the bases in our NS, and fit the movement with them. From Newton's Law, we can know that most of movement is formed with these bases. So, we use the regression with alterable bases to fit physical movement and get the analytic solution of it. With analytic solution, we can learn the movement pattern of object and also predict the trajectory of it.

Acknowledgement
This work is supported by Shenzhen Fundamental Research foundation (JCYJ20170818103918295) China Nature Science Foundation (Grant.No.21805234) and President's funds from CUHK-Shenzhen.